\documentclass[sigconf,nonacm]{acmart}

\usepackage[group-minimum-digits=4,per-mode=symbol]{siunitx}
\usepackage{subcaption}
\usepackage{booktabs}
\usepackage{enumitem}
\usepackage{tikz}
\usetikzlibrary{arrows.meta, positioning}

\usepackage[capitalize,noabbrev]{cleveref}
    \crefname{figure}{Fig.}{Figs.}
    \crefname{equation}{}{}
    \crefname{section}{Section}{Sections}
    \crefname{table}{Table}{Tables}

\graphicspath{{./figs/}}

\begin{document}

\title{Neuromorphic Parameter Estimation for Power Converter Health Monitoring Using Spiking Neural Networks}

\author{Hyeongmeen Baik}
\affiliation{%
  \institution{University of Wisconsin--Madison}
  \city{Madison}
  \state{Wisconsin}
  \country{USA}
}
\email{hyeongmeen.baik@wisc.edu}

\author{Hamed Poursiami}
\affiliation{%
  \institution{George Mason University}
  \city{Fairfax}
  \state{Virginia}
  \country{USA}
}
\email{hpoursia@gmu.edu}

\author{Maryam Parsa}
\affiliation{%
  \institution{George Mason University}
  \city{Fairfax}
  \state{Virginia}
  \country{USA}
}
\email{mparsa@gmu.edu}

\author{Jinia Roy}
\affiliation{%
  \institution{University of Wisconsin--Madison}
  \city{Madison}
  \state{Wisconsin}
  \country{USA}
}
\email{jinia.roy@wisc.edu}

\begin{abstract}
Always-on converter health monitoring demands sub-\si{\milli\watt} edge inference, a regime inaccessible to GPU-based physics-informed neural networks. This work separates spiking temporal processing from physics enforcement: a three-layer leaky integrate-and-fire SNN estimates passive component parameters while a differentiable ODE solver provides physics-consistent training by decoupling the ODE physics loss from the unrolled spiking loop. On an EMI-corrupted synchronous buck converter benchmark, the SNN reduces lumped resistance error from $25.8\%$ to $10.2\%$ versus a feedforward baseline, within the $\pm 10\%$ manufacturing tolerance of passive components, at a projected ${\sim}270\times$ energy reduction on neuromorphic hardware. Persistent membrane states further enable degradation tracking and event-driven fault detection via a $+5.5$ percentage-point spike-rate jump at abrupt faults. With $93\%$ spike sparsity, the architecture is suited for always-on deployment on Intel Loihi~2 or BrainChip Akida.
\end{abstract}



\maketitle

\section{Introduction}
\label{sec:introduction}

Physics-informed neural networks (PINNs) have emerged as a promising tool for power-converter digital twins, enabling online identification of component parameters such as inductance~$L$, capacitance~$C$, and lumped series resistance~$R_s$ from short transient measurements~\cite{raissi2019pinn, zhao2024piml_predictive}. Recent extensions include physics-informed machine-learning frameworks for DC-DC parameter estimation~\cite{stiasny2022piml_dcdc, chen2025epinn} and physics-informed predictive maintenance~\cite{zhao2024piml_predictive}. However, PINN inference requires floating-point multiply-accumulate (MAC) operations on a GPU or cloud accelerator, making always-on, edge-located condition monitoring impractical for cost- and power-con\-strained converter systems.

Spiking neural networks (SNNs) are brain-inspired models that more closely emulate the temporal dynamics of biological neurons, processing information through discrete spike events rather than continuous-valued activations~\cite{maass1997networks, roy2019towards}. Neuron models such as the leaky integrate-and-fire (LIF) neuron accumulate input over time in a membrane potential and emit a spike when a threshold is reached, naturally capturing temporal correlations in sequential data. On neuromorphic hardware such as Intel Loihi~2~\cite{davies2021loihi2} or BrainChip Akida~\cite{brainchip2022akida}, SNN inference consumes $1$--$\SI{10}{\milli\watt}$, roughly three orders of magnitude less than GPU-based inference~\cite{blouw2019loihi_benchmark}. This power budget is compatible with the auxiliary supplies of DC-DC converters, enabling a digital twin to reside physically next to the power stage without cloud connectivity.

Recent work on neuromorphic computing at scale~\cite{neuromorphic_scale2025} underscores the growing maturity of spike-based processors for always-on edge workloads. However, a key challenge in combining SNNs with physics-informed training for power converters is that computing ordinary differential equation (ODE) residuals via automatic differentiation through unrolled spiking dynamics is computationally prohibitive and numerically unstable. This is addressed by \emph{separating} the SNN's role (temporal parameter estimation) from the physics enforcement (differentiable ODE solver). The SNN processes a sparse temporal waveform and outputs three scalar parameters; the ODE solver then produces predicted waveforms, and gradients flow cleanly from the reconstruction loss through the solver back into the SNN weights.

Beyond deployment efficiency, LIF neurons offer a temporal processing advantage: the membrane potential integrates information across timesteps, enabling the SNN to extract parameter-relevant features from noisy waveforms. Furthermore, persistent membrane states across successive monitoring cycles enable capabilities that have no analogue in feedforward architectures: spike-rate changes directly encode whether the converter's operating condition has shifted, providing a native event-driven fault signal.

This paper presents an SNN+ODE architecture for physics-con\-sis\-tent parameter estimation targeting neuromorphic edge deployment. Section~\ref{sec:background} introduces the converter model and LIF neuron formulation. Section~\ref{sec:proposed} describes the proposed architecture. Section~\ref{sec:results} evaluates estimation accuracy on a buck converter benchmark with structured EMI. Section~\ref{sec:efficiency} analyzes inference energy, and Section~\ref{sec:degradation} extends the approach to degradation monitoring and event-driven fault detection. Section~\ref{sec:conclusion} concludes.

\section{Converter Model and LIF Neurons}
\label{sec:background}

\begin{figure}[!t]
\centering
\includegraphics[width=0.65\linewidth]{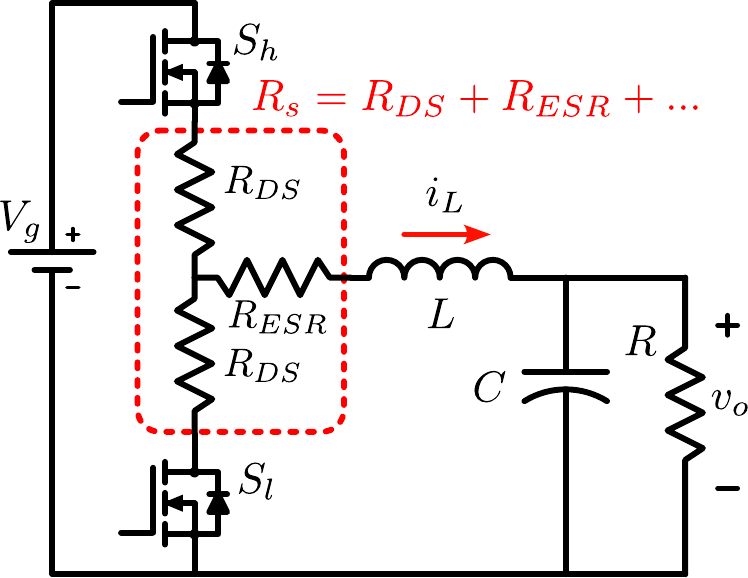}
\caption{Synchronous buck converter with lumped parasitic resistance $R_s$.}
\label{fig:schematic}
\end{figure}

\begin{figure}[!t]
\centering
\resizebox{0.85\linewidth}{!}{%
\begin{tikzpicture}[
    node distance=0.8cm,
    box/.style={draw, rounded corners=3pt, minimum height=0.9cm, minimum width=3.2cm, align=center, font=\small},
    arr/.style={-{Stealth[length=2.5mm]}, thick},
    garr/.style={-{Stealth[length=2.5mm]}, thick, red!70!black, dashed},
    lbl/.style={font=\scriptsize, midway},
]

\node[box, fill=teal!15] (input) {Noisy Waveform $i_L,\,V_o$};
\node[box, fill=teal!15, below=of input] (snn) {SNN Estimator (LIF$\times$3)};
\node[box, fill=teal!15, below=of snn] (params) {$\hat{L},\;\hat{C},\;\hat{R}_s$};
\node[box, fill=teal!15, below=of params] (ode) {Differentiable ODE Solver};
\node[box, fill=teal!15, below=of ode] (pred) {Predicted $\hat{i}_L,\,\hat{V}_o$};
\node[box, fill=teal!15, below=of pred] (loss) {Waveform Loss $\mathcal{L}$};

\draw[arr] (input) -- node[lbl, right] {100 spike-encoded timesteps} (snn);
\draw[arr] (snn) -- node[lbl, right] {membrane readout $\to$ $e^{\log\theta}$} (params);
\draw[arr] (params) -- node[lbl, right] {RK4 integration of Eqs.\,(1)--(2)} (ode);
\draw[arr] (ode) -- node[lbl, right] {300 predicted time points} (pred);
\draw[arr] (pred) -- node[lbl, right] {MSE vs.\ measured waveform} (loss);

\draw[garr] (loss.west) -- ++(-1.2,0) |- node[lbl, left, pos=0.25] {Backpropagation} (snn.west);

\end{tikzpicture}%
}
\caption{SNN+ODE architecture. Solid arrows: forward inference path. Dashed arrow: backpropagation through the ODE solver into SNN weights.}
\label{fig:architecture}
\end{figure}
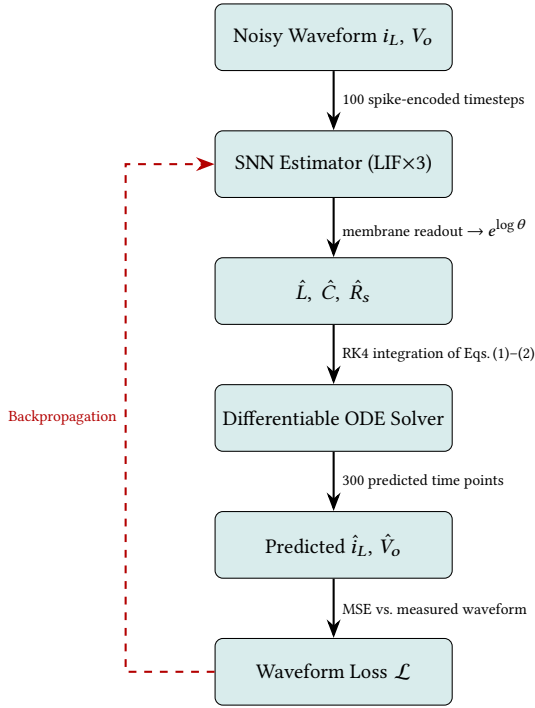

Consider the synchronous buck converter shown in Fig.~\ref{fig:schematic}, with averaged-model dynamics:
\begin{align}
    L \frac{di_L}{dt} &= d\,V_g - V_o - R_s\,i_L, \label{eq:ode1} \\
    C \frac{dV_o}{dt} &= i_L - \frac{V_o}{R}, \label{eq:ode2}
\end{align}
where $d$ is the duty ratio, $V_g$ the input voltage, $R$ the load resistance, and $R_s$ is the lumped series resistance often aggregating MOSFET on-state resistance ($R_{DS}$), inductor equivalent series resistance ($R_{ESR}$), and trace losses. A PINN approximates the state trajectories $\hat{i}_L(t)$ and $\hat{V}_o(t)$ with a neural network $f_\theta(t)$ and embeds \cref{eq:ode1,eq:ode2} as soft constraints via the training loss, driving the estimated parameters toward their true values by minimizing the ODE residuals at collocation points.

In practice, converter measurements are corrupted by switching-synchronous electromagnetic interference (EMI): Gaussian-en\-ve\-lope spikes near pulse-width modulation (PWM) edges superimposed on broadband Gaussian background noise. These heavy-tailed, time-correlated artifacts bias standard mean squared error (MSE) based parameter estimation, particularly for $R_s$, which has the weakest observable signature in the waveforms. This motivates the use of a temporal processing architecture that can integrate information across timesteps to separate the slow parameter-dependent dynamics from fast noise transients.

LIF neurons provide such temporal integration. A LIF neuron updates its membrane potential $U$ at each discrete timestep~$k$:
\begin{equation}
    U[k] = \beta\,U[k-1] + W\,S_\mathrm{in}[k] - S_\mathrm{out}[k-1]\,U_\mathrm{thr},
    \label{eq:lif}
\end{equation}
where $\beta \in (0,1)$ is the membrane decay factor, $W$ is the synaptic weight, $S_\mathrm{in}$ is the input spike (or current), $U_\mathrm{thr}$ is the firing threshold, and $S_\mathrm{out}[k] = \Theta(U[k] - U_\mathrm{thr})$ is the output spike ($\Theta$ is the Heaviside function). The subtract-reset mechanism $U \leftarrow U - U_\mathrm{thr}$ preserves residual membrane charge after firing. For gradient-based training, the non-differentiable $\Theta$ is replaced by a surrogate gradient~\cite{neftci2019surrogate}; the fast sigmoid~\cite{zenke2021remarkable} is adopted:
\begin{equation}
    \tilde{\sigma}'(x) = \frac{1}{(1 + \alpha|x|)^2},
    \label{eq:surrogate}
\end{equation}
with slope $\alpha = 25$, enabling standard backpropagation through unrolled spiking dynamics using snnTorch~\cite{eshraghian2023snntorch}.
\section{Proposed SNN+ODE Architecture}
\label{sec:proposed}

\begin{figure}[!t]
\centering
\includegraphics[width=\linewidth]{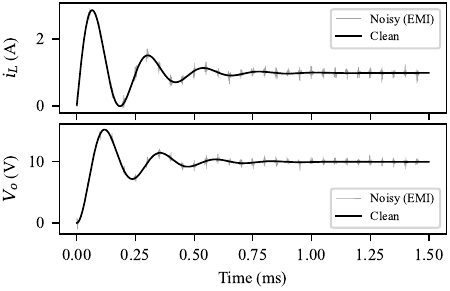}
\caption{Buck converter waveforms under structured EMI noise.}
\label{fig:emi_noise}
\end{figure}

A naive approach to combining SNNs with physics-informed training would embed spiking layers inside a PINN and compute ODE residuals via automatic differentiation through the unrolled spiking loop. This is problematic for two reasons: (1)~training is significantly slower than feedforward PINNs due to the computational graph spanning every timestep, and (2)~surrogate gradients through deep temporal unrolling cause noisy or vanishing derivatives, particularly for the current equation $di_L/dt$.

Both issues are resolved by \emph{separating} the SNN's role from the physics enforcement:
\begin{itemize}[leftmargin=*,topsep=2pt,itemsep=1pt]
    \item \textbf{SNN Estimator}: processes the noisy waveform temporally and outputs three scalar parameters ($L$, $C$, $R_s$).
    \item \textbf{Differentiable ODE Solver}: takes the estimated parameters and integrates \cref{eq:ode1,eq:ode2} to produce predicted waveforms.
    \item \textbf{Reconstruction Loss}: MSE between predicted and measured waveforms.
\end{itemize}
Gradients flow from $\mathcal{L}$ through the ODE solver into $(L, C, R_s)$ and then into SNN weights via surrogate gradients, but the ODE physics enforcement itself does not require differentiating through the unrolled spiking loop.

\begin{table}[!t]
\centering
\caption{Parameter identification results (best checkpoint, 3{,}000 epochs).}
\label{tab:results}
\scalebox{0.85}{
\begin{tabular}{lccccc}
\toprule
 & True & \multicolumn{2}{c}{FF+ODE} & \multicolumn{2}{c}{SNN+ODE} \\
\cmidrule(lr){3-4} \cmidrule(lr){5-6}
 & & Value & Error & Value & Error \\
\midrule
$L$ ($\si{\micro\henry}$) & 138 & 117.1 & 15.2\% & 115.3 & 16.4\% \\
$C$ ($\si{\micro\farad}$) & 10.0 & 11.0 & 10.1\% & 10.9 & \textbf{8.8\%} \\
$R_s$ ($\si{\ohm}$)       & 0.100 & 0.074 & 25.8\% & 0.090 & \textbf{10.2\%} \\
\bottomrule
\end{tabular}
}
\end{table}

\subsection{SNN Estimator}

The SNN estimator receives a subsampled measured waveform $\mathbf{x}[k] = [i_L(t_k),\, V_o(t_k)]$ and processes each sample as one SNN timestep. As shown in Fig.~\ref{fig:architecture}, the architecture consists of three LIF hidden layers with subtract-reset (\cref{eq:lif}), followed by a leaky integrator (LI) readout layer that accumulates spike-driven currents into a smooth membrane state without resetting. After all $N_s$ timesteps, the final membrane state yields the estimated parameters in log space:
\begin{equation}
    [\log L,\; \log C,\; \log R_s] = \mathbf{m}_\mathrm{out}[N_s] + \mathbf{b}_\mathrm{param},
    \label{eq:output}
\end{equation}
where $\mathbf{b}_\mathrm{param}$ is a learned bias initialized near the expected parameter range. The physical parameters are recovered as $L = e^{\log L}$, etc.

\textbf{Temporal integration advantage.} Unlike a feedforward network that receives a flattened snapshot of all measurement points, the SNN processes the waveform sample-by-sample. The LIF membrane potential acts as a recurrent state that integrates information across time, enabling the network to track transient dynamics (startup overshoot, oscillation frequency, damping) that encode the physical parameters. This temporal processing extracts more information from the same number of samples, particularly for $R_s$, whose effect appears as a slow damping modulation rather than a localized waveform feature.

\begin{figure}[!t]
\centering
\includegraphics[width=\linewidth]{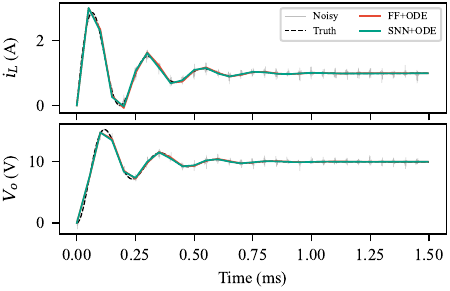}
\caption{Waveform reconstruction from estimated parameters.}
\label{fig:waveforms}
\vspace{-5pt}
\end{figure}

\subsection{Differentiable ODE Solver}

Given the estimated parameters $(\hat{L}, \hat{C}, \hat{R}_s)$, \cref{eq:ode1,eq:ode2} are integrated from zero initial conditions using a 4th-order Runge-Kutta (RK4) solver (torchdiffeq~\cite{chen2018neuralode}) to obtain predicted waveforms $[\hat{i}_L(t),\, \hat{V}_o(t)]$.

The reconstruction loss is:
\begin{equation}
    \mathcal{L} = \frac{1}{\sigma_{i_L}^2}\,\text{MSE}(\hat{i}_L, i_L^\mathrm{meas}) + \frac{1}{\sigma_{V_o}^2}\,\text{MSE}(\hat{V}_o, V_o^\mathrm{meas}),
    \label{eq:loss}
\end{equation}
where $i_L^\mathrm{meas}$ and $V_o^\mathrm{meas}$ are the measured waveforms, and $\sigma_{i_L}$, $\sigma_{V_o}$ are their temporal standard deviations, normalizing the two terms to comparable scale.


\begin{figure*}[!t]
\centering
\begin{subfigure}[t]{0.65\textwidth}
  \centering
  \includegraphics[width=\linewidth]{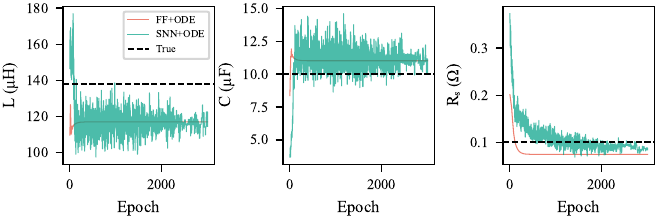}
  \caption{}
  \label{fig:param_conv}
\end{subfigure}
\hfill
\begin{subfigure}[t]{0.34\textwidth}
  \centering
  \includegraphics[width=\linewidth]{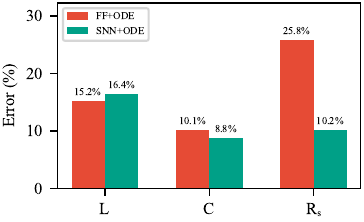}
  \caption{}
  \label{fig:error_bar}
\end{subfigure}
\caption{Training results: (a) parameter convergence over 3{,}000 epochs, (b) final identification errors at best checkpoint.}
\label{fig:convergence}
\end{figure*}

\begin{figure*}[!t]
\centering
\begin{subfigure}[t]{0.24\textwidth}
  \centering
  \includegraphics[width=\linewidth]{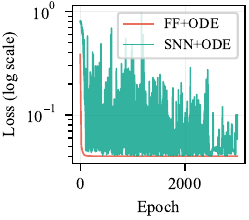}
  \caption{}
  \label{fig:loss}
\end{subfigure}
\hfill
\begin{subfigure}[t]{0.74\textwidth}
  \centering
  \includegraphics[width=\linewidth]{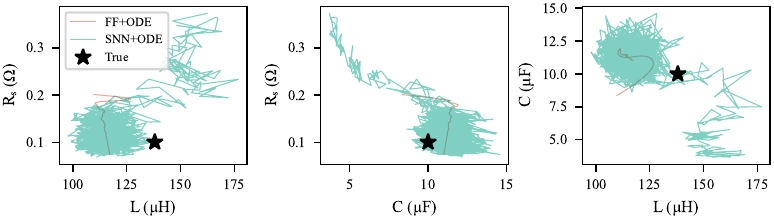}
  \caption{}
  \label{fig:trajectory_3d}
\end{subfigure}
\caption{Optimization dynamics: (a) training loss curves, (b) parameter trajectories in $(L, C, R_s)$ space.}
\label{fig:training}
\end{figure*}

\section{Simulation Validation}
\label{sec:results}

The benchmark uses a synchronous buck converter ($V_g = \SI{20}{\volt}$, $d = 0.5$, $f_s = \SI{10}{\kilo\hertz}$, $R = \SI{10}{\ohm}$) with true parameters $L = \SI{138}{\micro\henry}$, $C = \SI{10}{\micro\farad}$, $R_s = \SI{0.1}{\ohm}$. The averaged ODE (\cref{eq:ode1,eq:ode2}) is simulated over $T_\mathrm{sim} = \SI{1.5}{\milli\second}$ (15~switching cycles) with $\Delta t = \SI{0.5}{\micro\second}$, yielding 3{,}000 time points per state variable.

Structured EMI noise is added (see Fig.~\ref{fig:emi_noise}): Gaussian background ($\sigma = 2\%$ of signal standard deviation) plus switching-edge spikes modeled as Gaussian-envelope pulses ($\SI{4}{\micro\second}$ width, $25\%$ peak amplitude) at each turn-on and turn-off edge.

Both models receive $N_s = 100$ subsampled waveform points (stride~30, $\Delta t_\mathrm{SNN} = \SI{15}{\micro\second}$), chosen to limit the SNN unrolling depth while retaining the key transient features. The SNN estimator uses hidden dimension $H = 128$, membrane decay $\beta = 0.9$, LI readout decay $\beta_\mathrm{out} = 0.95$, and the $\mathbf{b}_\mathrm{param}$ bias receives a $3\times$ learning rate, as it must traverse a larger distance in log-space to reach the target parameter region. The ODE solver uses RK4 integration with step size $\SI{50}{\micro\second}$, producing 300 predicted time points over $T_\mathrm{sim} = \SI{1.5}{\milli\second}$ for the reconstruction loss; a finer resolution than the SNN input is used here because the loss computation benefits from denser sampling while forward integration is inexpensive. The feedforward (FF) estimator receives the same 100 points as a flattened 200-dimensional input vector and processes it through three Tanh hidden layers of 128~neurons. Both models are trained with Adam ($\text{lr} = 5 \times 10^{-5}$) and cosine annealing over 3{,}000 epochs, with gradient clipping set to $0.3$ for the SNN to suppress noisy surrogate-gradient spikes. The best model checkpoint (lowest reconstruction loss) is retained for evaluation. Results for both architectures are compared in Table~\ref{tab:results}.

\textbf{$R_s$ recovery.} The SNN+ODE achieves $10.2\%$ $R_s$ error versus $25.8\%$ for the feedforward baseline, i.e., a $2.5\times$ improvement. Both models use identical MSE loss and receive the same 100-point noisy waveform. The SNN's advantage comes from its LIF temporal integration: rather than processing a flattened snapshot, the SNN tracks the waveform's damping dynamics across timesteps, extracting a more accurate $R_s$ estimate from the time-dependent energy dissipation signature.

\textbf{$C$ identification.} The SNN also outperforms the FF baseline on capacitance, i.e., $8.8\%$ vs $10.1\%$, likely because the membrane potential integrates oscillation-period information that directly encodes the $LC$ resonance.

\textbf{$L$ identification.} Both models show similar performance on $L$ (${\sim}15$--$16\%$). This parameter is primarily determined by the startup ramp rate, which is well-captured by either architecture but remains sensitive to EMI corruption of the initial transient.

\begin{table}[!t]
\centering
\caption{Energy-per-inference comparison using published hardware measurements.}
\label{tab:efficiency}
\scalebox{0.82}{
\begin{tabular}{lccc}
\toprule
 & FF+ODE & SNN+ODE & Source \\
 & (Cortex-M) & (Neuromorphic) & \\
\midrule
Operations & 58{,}752 MACs & 333{,}470 SOPs & Measured \\
Energy/op & ${\sim}$15\,nJ/MAC & 9.9\,pJ/SOP & \cite{lai2018cmsis, huang2023neuromorphic} \\
Spike sparsity & --- & 92.9\% & Measured \\
\textbf{Energy/inference} & ${\sim}$\textbf{881\,\si{\micro\joule}} & ${\sim}$\textbf{3.3\,\si{\micro\joule}} & Estimated \\
\textbf{Ratio} & \multicolumn{2}{c}{${\sim}270\times$ reduction} & \\
\bottomrule
\end{tabular}
}
\vspace{-10pt}
\end{table}

The SNN's estimation errors for $R_s$ and $C$ fall within the typical $\pm 10\%$ manufacturing tolerance of passive components, suggesting that the accuracy is sufficient for detecting out-of-tolerance degradation rather than absolute parameter identification. The waveform reconstruction from the estimated parameters is shown in Fig.~\ref{fig:waveforms}.

\textbf{Training dynamics.} As shown in Fig.~\ref{fig:param_conv} and Fig.~\ref{fig:error_bar}, the SNN parameter estimates oscillate before settling due to noisy surrogate gradients, motivating the use of best-checkpoint selection. The SNN loss curve (see Fig.~\ref{fig:loss}) converges more slowly than FF but reaches a comparable final value. Fig.~\ref{fig:trajectory_3d} visualizes the optimization paths in parameter space.

\vspace{-5pt}            
\section{Inference Efficiency Analysis}
\label{sec:efficiency}

A key advantage of SNNs is that computation occurs only when neurons fire. During inference on the trained SNN estimator, all spike activations across hidden layers are recorded and the following is computed:
\begin{equation}
    \text{Sparsity} = 1 - \frac{\sum_{k,l} S_l[k]}{N_s \times \sum_l H_l},
    \label{eq:sparsity}
\end{equation}
where $S_l[k]$ is the number of spikes in layer~$l$ at timestep~$k$, $N_s = 100$ is the number of timesteps, and $H_l$ is the neuron count in layer~$l$. The measured sparsity is $92.9\%$, meaning only $7.1\%$ of possible spike events actually occur.

For the feedforward estimator, inference cost is measured in multiply-accumulate operations (MACs):
\begin{equation}
    \text{MACs}_\mathrm{FF} = 200{\times}128 + 2{\times}128{\times}128 + 128{\times}3 = 58{,}752.
\end{equation}

For the SNN estimator, the relevant metric is synaptic operations (SOPs), where a synapse is activated only when its presynaptic neuron fires:
\begin{equation}
    \text{SOPs}_\mathrm{SNN} = \sum_{k=1}^{N_s} \sum_{l} \| S_{l-1}[k] \|_0 \times H_l,
    \label{eq:sops}
\end{equation}
where $\| S_{l-1}[k] \|_0$ counts the number of active (spiking) neurons at timestep~$k$. On the benchmark signal, the SOP count is 333{,}470, which exceeds the FF MAC count in raw operations. However, the energy per operation differs by over an order of magnitude: each SOP is an \emph{accumulate-only} operation (binary spike $\times$ weight = add weight), whereas each MAC involves a full multiply-accumulate on multi-bit operands.

Table~\ref{tab:efficiency} provides an energy-per-inference estimate using published per-operation costs from fabricated hardware. For the SNN, a measured energy of $\SI{9.9}{\pico\joule}$/SOP from a 40-nm neuromorphic processor~\cite{huang2023neuromorphic} is adopted, yielding $333{,}470 \times 9.9\,\si{\pico\joule} \approx \SI{3.3}{\micro\joule}$ per inference. For a more conservative estimate on Intel Loihi~2, published benchmarks report 3--$\SI{6}{\micro\joule}$ per inference for comparably-sized SNNs~\cite{davies2018loihi}. For the FF baseline, deployment on an ARM Cortex-M class microcontroller (MCU) with optimized NN kernels~\cite{lai2018cmsis} yields ${\sim}\SI{881}{\micro\joule}$ per inference, resulting in an estimated ${\sim}270\times$ energy reduction.

\begin{figure}[!t]
\centering
\includegraphics[width=0.9\linewidth]{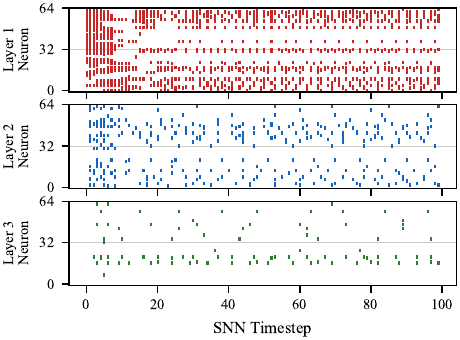}
\caption{Spike raster plot (first 64 of 128 neurons shown per layer, 100 timesteps).}
\label{fig:spike_raster}

\centering
\includegraphics[width=0.9\linewidth]{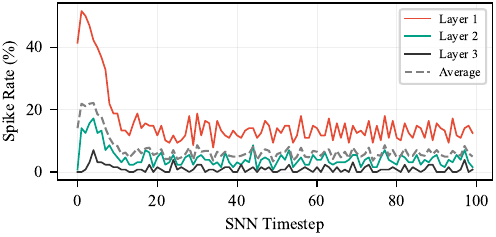}
\caption{Per-layer spike rate over SNN timesteps.}
\label{fig:spike_rate}
\vspace{-5pt}
\end{figure}

The spike rates exhibit a hierarchical sparsity pattern (see Fig.~\ref{fig:spike_raster} and Fig.~\ref{fig:spike_rate}): Layer~1 fires at $15.8\%$, Layer~2 at $4.5\%$, and Layer~3 at $1.1\%$. The decreasing activity deeper in the network indicates progressively more abstract and sparser encoding. As shown in Fig.~\ref{fig:spike_rate}, spike activity peaks during the startup transient (timesteps 0--30) and decreases during steady state, consistent with higher information content in the transient region.

At ${\sim}\SI{3.3}{\micro\joule}$ per inference and a monitoring rate of one snapshot per second, the SNN consumes ${\sim}\SI{3.3}{\micro\watt}$, well within the auxiliary power budget of DC-DC converters. The SNN architecture is directly compatible with existing neuromorphic deployment tools: snnTorch models can be exported to BrainChip MetaTF for Akida deployment or converted via Lava for Intel Loihi~2.

\begin{figure*}[!t]
\centering
\includegraphics[width=0.9\linewidth]{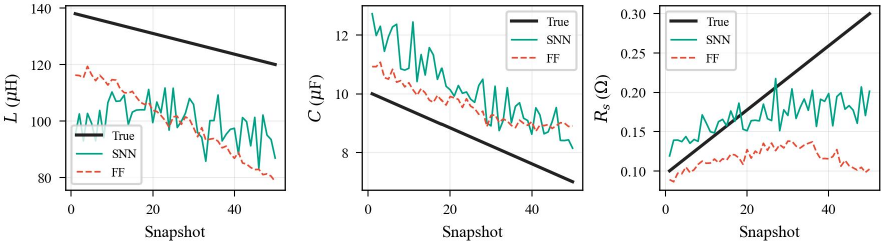}
\caption{All three parameters tracked during simultaneous degradation.}
\label{fig:all_tracking}
\end{figure*}

\begin{figure}[!t]
\centering
\includegraphics[width=0.9\linewidth]{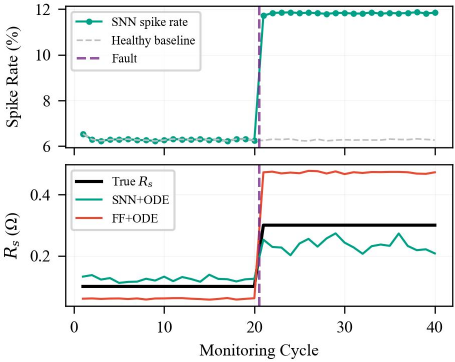}
\caption{Abrupt fault detection via persistent membrane states.}
\label{fig:fault_spike}
\end{figure}

\begin{figure}[!t]
\centering
\includegraphics[width=0.9\linewidth]{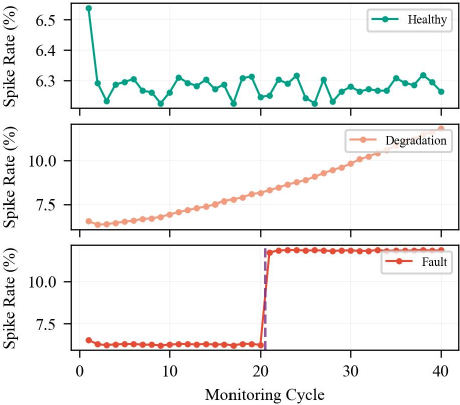}
\caption{Spike rate under three monitoring scenarios.}
\label{fig:paradigm_three}
\end{figure}

\section{Degradation Monitoring and Fault Detection}
\label{sec:degradation}

SNNs have shown promise for regression~\cite{hauser2022snn_regression}, anomaly detection~\cite{dampfhoffer2023snn_anomaly}, battery fault diagnosis~\cite{li2025snn_battery}, and vibration-based predictive maintenance on neuromorphic hardware~\cite{snn_predictive_maint2025, neuro_bearing_cm2023}. To evaluate this capability for converter health monitoring, the single-operating-point estimator is extended to a multi-condition setting.

Both the SNN and FF estimators are retrained on 200 waveforms generated with randomly sampled parameters: $L \in [80, 200]\,\si{\micro\henry}$, $C \in [5, 15]\,\si{\micro\farad}$, $R_s \in [0.02, 0.5]\,\si{\ohm}$, for 6{,}000 epochs with $\text{lr} = 10^{-4}$. Gradual component aging is then simulated over 50 monitoring snapshots: $R_s$: $0.1 \to 0.3\,\si{\ohm}$, $C$: $10 \to 7\,\si{\micro\farad}$, $L$: $138 \to 120\,\si{\micro\henry}$. Each snapshot uses an independently generated noisy waveform; both models are evaluated in inference mode.

Table~\ref{tab:degradation} reports the mean estimation error. The SNN achieves $1.9\times$ lower $R_s$ tracking error than the FF baseline, i.e., $19.1\%$ vs $36.8\%$. As shown in Fig.~\ref{fig:all_tracking}, the SNN's $R_s$ estimate follows the upward degradation trend, while the FF estimate remains nearly flat because the FF network has memorized a single parameter region and cannot track drift.

\begin{table}[!t]
\centering
\caption{Mean parameter estimation error during degradation (50 snapshots).}
\label{tab:degradation}
\scalebox{0.85}{
\begin{tabular}{lccc}
\toprule
 & $L$ error & $C$ error & $R_s$ error \\
\midrule
FF+ODE  & 22.8\% & 13.3\% & 36.8\% \\
SNN+ODE & 21.5\% & 13.4\% & \textbf{19.1\%} \\
\bottomrule
\end{tabular}
}
\end{table}


A unique advantage of SNNs over feedforward networks is the ability to maintain \emph{persistent membrane states} across successive monitoring cycles. Here, membrane states are carried from one monitoring snapshot to the next, so that the LIF neurons accumulate temporal context over long time horizons.

Three scenarios are tested over 40 monitoring cycles, each using a 1.5\,ms waveform snapshot (100 SNN timesteps):
\begin{itemize}[leftmargin=*,topsep=2pt,itemsep=1pt]
    \item \textbf{Healthy baseline}: $R_s = 0.1\,\si{\ohm}$ constant.
    \item \textbf{Abrupt fault}: $R_s$ jumps from $0.1$ to $0.3\,\si{\ohm}$ at cycle~21.
    \item \textbf{Gradual degradation}: $R_s$ ramps linearly from $0.1$ to $0.3\,\si{\ohm}$.
\end{itemize}

Table~\ref{tab:event-driven} and Fig.~\ref{fig:fault_spike} summarize the results. In the healthy baseline, the spike rate stabilizes at $6.3\%$ as membranes reach equilibrium. When an abrupt fault occurs, the spike rate jumps from $6.3\%$ to $11.8\%$ at the exact transition cycle, a $+5.5$ percentage-point increase that serves as an immediate fault indicator. During gradual degradation, spike rate rises from $6.6\%$ to $10.9\%$, tracking the $R_s$ drift.

As shown in Table~\ref{tab:event-driven}, the SNN achieves lower $R_s$ mean absolute error (MAE) than the FF baseline in both pre-fault and post-fault regimes. In the abrupt fault scenario, the post-fault MAE is $0.067\,\si{\ohm}$ for the SNN versus $0.172\,\si{\ohm}$ for the FF baseline, a $2.6\times$ improvement. Beyond estimation accuracy, the spike-rate jump of $+5.5$ percentage points provides an immediate fault indicator that has no analogue in the feedforward architecture. On neuromorphic hardware, spike activity directly determines energy consumption: when membranes are stable (healthy), computation is minimal; when a fault disrupts the membrane equilibrium, the resulting spike burst alerts the system. Fig.~\ref{fig:paradigm_three} compares the spike rate across all three scenarios.

\begin{table}[!t]
\centering
\caption{Event-driven monitoring with persistent membrane states.}
\label{tab:event-driven}
\scalebox{0.82}{
\begin{tabular}{lcccc}
\toprule
Scenario & \multicolumn{2}{c}{Spike Rate} & \multicolumn{2}{c}{$R_s$ MAE (\si{\ohm})} \\
\cmidrule(lr){2-3} \cmidrule(lr){4-5}
 & Early & Late & SNN & FF \\
\midrule
Healthy            & 6.3\% & 6.3\%  & ---              & ---   \\
Abrupt (pre-fault) & 6.3\% & ---    & \textbf{0.024}   & 0.039 \\
Abrupt (post-fault)& ---   & 11.8\% & \textbf{0.067}   & 0.172 \\
Gradual deg.       & 6.6\% & 10.9\% & \textbf{0.053}   & 0.099 \\
\bottomrule
\end{tabular}
}
\vspace{-5pt}
\end{table}

\vspace{-5pt}            
\section{Conclusion}
\label{sec:conclusion}

This work presented an SNN+ODE architecture for physics-con\-sis\-tent converter parameter estimation targeting always-on, edge-located health monitoring on neuromorphic hardware. By separating spiking temporal processing from ODE-based physics enforcement, the architecture decouples the ODE physics loss from the unrolled spiking loop while preserving physics consistency, and produces models directly deployable on commercial neuromorphic platforms.

Using only spike-based computation, the SNN achieves $R_s$ error of $10.2\%$ and $C$ error of $8.8\%$ on EMI-corrupted waveforms, at an estimated ${\sim}\SI{3.3}{\micro\joule}$ per inference, ${\sim}270\times$ lower than a feedforward baseline on an ARM Cortex-M edge processor. This energy footprint allows the SNN to operate continuously within the auxiliary power budget of a DC-DC converter, enabling a monitoring paradigm where the estimator resides physically next to the power stage. Persistent membrane states across monitoring cycles provide event-driven fault detection: spike rate jumps $+5.5$ percentage points at abrupt $R_s$ faults, serving as a native trigger signal without explicit threshold logic. For gradual degradation, the SNN's temporal integration tracks $R_s$ drift with $19.1\%$ mean error versus $36.8\%$ for the feedforward baseline.

The approach involves inherent trade-offs. Feedforward PINNs with noise-aware loss design and higher-precision floating-point computation can achieve lower $R_s$ errors, but require GPU-class hardware consuming orders of magnitude more power. SNN training exhibits surrogate gradient noise, requiring best-checkpoint selection. These are acceptable costs for applications where the deployment power budget, not training-time accuracy, is the binding constraint.

Future work includes multi-trial statistical validation across diverse operating conditions and noise realizations; combining spiking dynamics with robust loss functions to narrow the accuracy gap; training with persistent membrane states across consecutive waveforms for inter-cycle fault adaptation; and hardware deployment on BrainChip Akida or Intel Loihi~2 for measured power and latency characterization.

\bibliographystyle{ACM-Reference-Format}
\bibliography{bibliography}

\end{document}